\pdfoutput=1

\documentclass[11pt]{article}

\usepackage[]{acl}

\usepackage{times}
\usepackage{latexsym}

\usepackage[T1]{fontenc}

\usepackage[utf8]{inputenc}

\usepackage{microtype}

\usepackage{graphicx} 
\usepackage{caption} 
\usepackage{algorithm}
\usepackage{algorithmic}
\usepackage{multirow} %
\usepackage{makecell}
\usepackage{amsmath}%
\usepackage{MnSymbol}%
\usepackage{wasysym}%
\usepackage{lipsum,subcaption}
\usepackage{newfloat}
\usepackage{listings}
\usepackage{dsfont}

\usepackage{tabularx}
\usepackage{tikz}
\usetikzlibrary{shapes.misc,shadows}

\title{Multimodal Dialogue Response Generation
}

\author{Qingfeng Sun$^1$ \quad Yujing Wang$^2$ \quad Can Xu$^1$ \quad  Kai Zheng$^1$ \quad Yaming Yang$^2$ \\ \quad {\bf Huang Hu}$^1$  \quad {\bf Fei Xu}$^1$ \quad {\bf Jessica Zhang}$^1$ \quad {\bf Xiubo Geng}$^1$ \quad {\bf Daxin Jiang}$^1$\thanks{\quad Corresponding author.}  \\
      $^1$Microsoft STC Asia \quad $^2$Microsoft Research Asia\\ 
      \texttt{\{qins,yujwang,caxu,zhengkai,yayaming,huahu,fexu,}\\
      \texttt{jessicaz,xigeng,djiang\}@microsoft.com}}

\begin{document}
\maketitle
\begin{abstract}

Responsing with image has been recognized as an important capability for an intelligent conversational agent. Yet existing works only focus on exploring the multimodal dialogue models which depend on retrieval-based methods, but neglecting generation methods. To fill in the gaps, we first present a new task: multimodal dialogue response generation (MDRG) - given the dialogue context, one model needs to generate a text or an image as response. Learning such a MDRG model often requires multimodal dialogues containing both texts and images which are difficult to obtain. Motivated by the challenge in practice, we consider MDRG under a natural assumption that only limited training examples are available. Under  such a low-resource setting, we devise a novel conversational agent, Divter, in order to isolate parameters that depend on multimodal dialogues from the entire generation model. By this means, the major part of the model can be learned from a large number of text-only dialogues and text-image pairs respectively, then the whole parameters can be well fitted using just a few training examples. Extensive experiments demonstrate our method achieves state-of-the-art results in both automatic and human evaluation, and can generate informative text and  high-resolution image responses.

\end{abstract}

\section{Introduction}

With the development of instant messaging technology in the recent decades, the intermediary of online conversation has also changed from pure text to a variety of visual modalities (e.g., image, gif animation, short video). Similar to communicating by the messenger tools (e.g., Facebook, WhatsApp, WeChat) in reality, an excellent intelligent conversational agent should not only be able to converse freely with plain text, but also have the ability to perceive and share the real visual physical world.

\begin{figure}[!htb]
\centering
     \includegraphics[width=0.5\textwidth, scale=1, trim=346 169 333 35,clip]{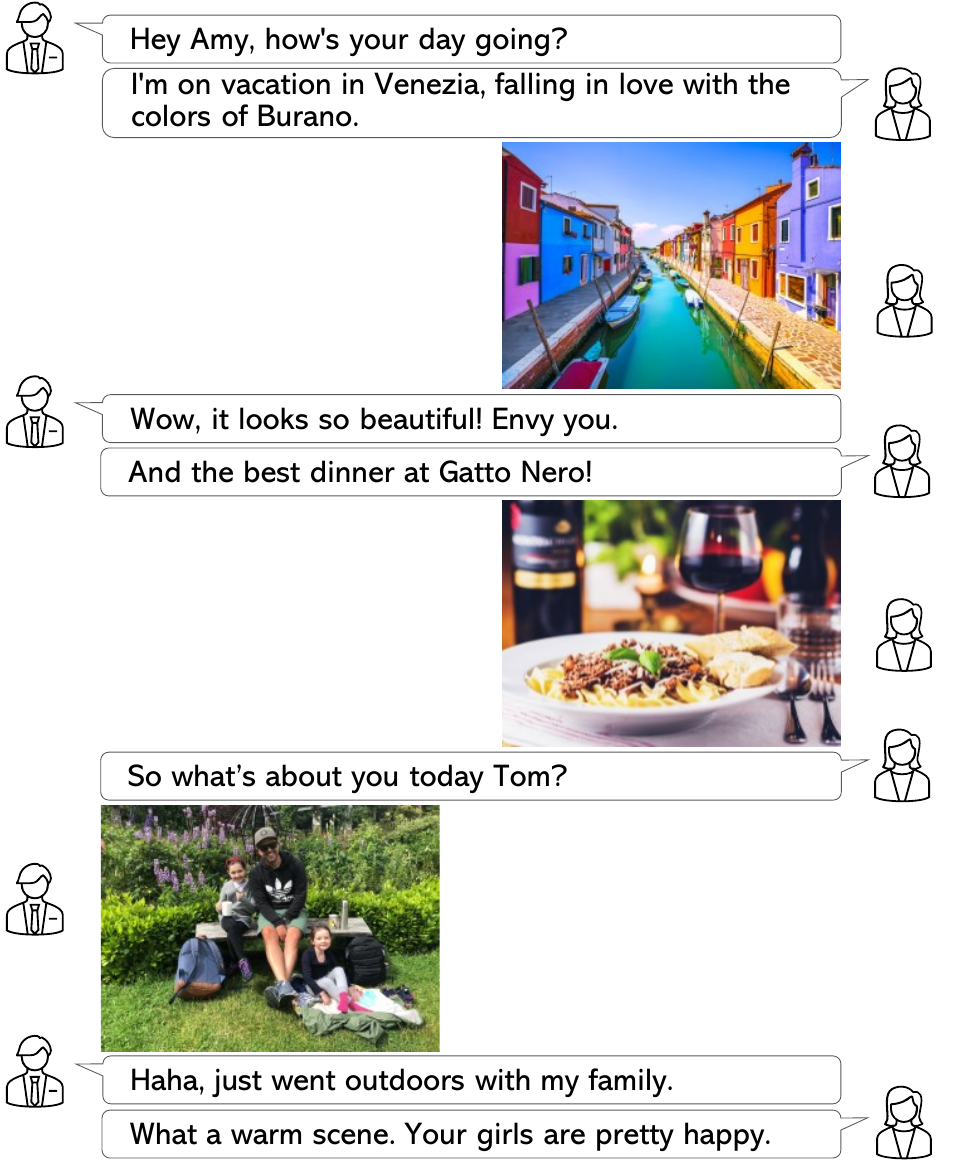}
     \caption{An example of human conversations. They are talking about vacation and outdoors with both text and various images.}
     \label{fig:intro_case}
\end{figure}

\begin{figure*}[bht]
\centering
     \includegraphics[scale=2, width=1\textwidth,  trim=58 145 78 120,clip]{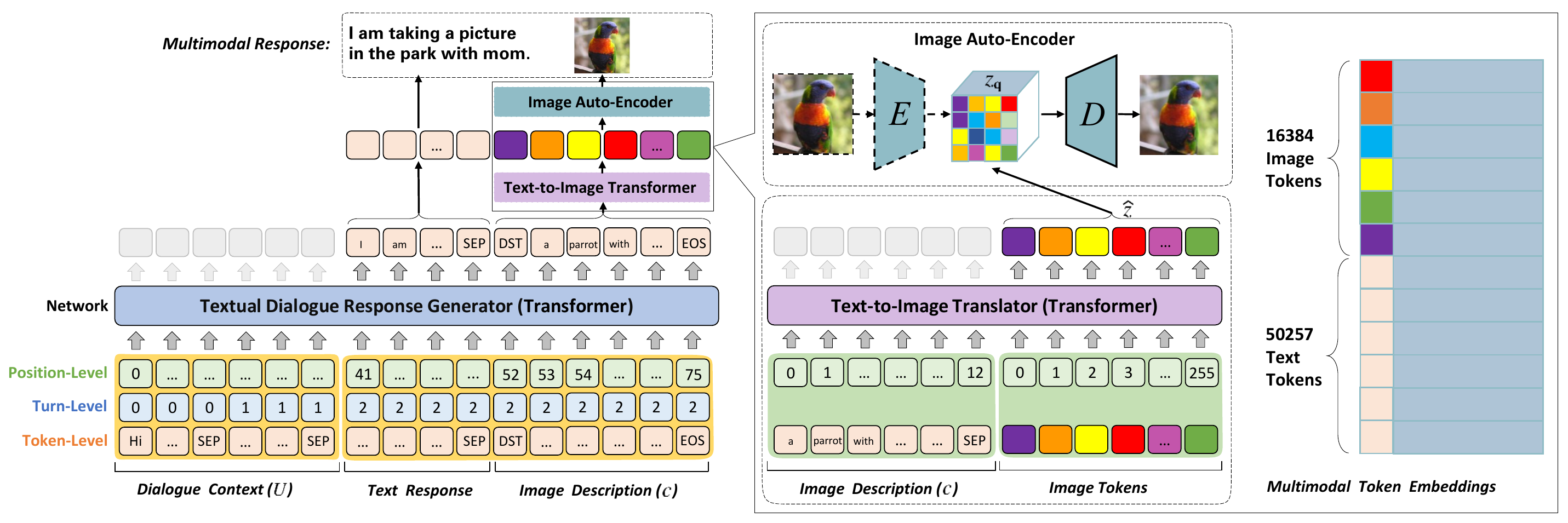}
     \caption{The overview of our multimodal dialogue response generation model. The Textual Dialogue Response Generator takes the text dialogue context $U$ as input and generates a sequence contains text response and a image description (e.g., ``a parrot with red belly and green back is standing on the railing.''). With the description as a condition, the Text-to-Image Translator generates image  representation $\hat{z}$. The Image Decoder $\mathcal{V}_{D}$ reconstructs $\hat{z}$ to a realistic and consistent high resolution image.}
     \label{fig:model}
\end{figure*}

Although recently some large-scale pre-trained text-only dialogue generation models, such as DialoGPT \cite{zhang2019dialogpt}, Blender \cite{roller-etal-2021-recipes}, Meena \cite{adiwardana2020humanlike}, have shown excellent performance, they still cannot rely exclusively on plain text to completely simulate the rich experience of visual perception. Recently, various vision-language tasks have been introduced and attracted widespread attention, such as visual question answering \cite{Exploring2015,Hierarchical2015,Anderson2017up-down,li2019relation,huang-etal-2020-aligned}, image captioning \cite{xu2016show,spice2016,ghanimifard-dobnik-2019-goes,cornia2020m2}, image-grounded dialogue \cite{das2017visual,Yang_Wu_Hu_Xu_Wang_Li_2021,agarwal-etal-2020-history,qi2019two,chen-etal-2021-multimodal, liang-etal-2021-maria}. 

Specifically, in human conversations, the images can easily show rich visual perception, which is hard to be expressed by plain text. As the example shown in Figure \ref{fig:intro_case}, images are required in at least three circumstances: (i) the other speaker has little knowledge (e.g., colorful Burano, in the 1st image) of the objects only you had seen; (ii) to share more details (e.g., red wine and pasta, in the 2nd image) of the objects even you have common knowledge of them; (iii) to express your emotions (e.g., happy, in the 3rd image) about a specific event. An existing related task is photo sharing \cite{zang-etal-2021-photochat}, which aims to select and share the image based on the textual context, is a challenging task that requires models to understand the background story which complemented by human imaginations, rather than to locate related visual objects or explicitly mention main visible content in the image as the previous works do. \citet{zang-etal-2021-photochat} propose a retrieval-based method to resolve the above challenge. However, the performance of the retrieval-based method is limited in specific domains by the size of the pre-constructed conversational history repository, especially for long-tail contexts that are not covered in the history, where the set of image responses of a retrieval system is also fixed. On the other hand, a better way is to generate a new one accordingly. 



In this paper, we formulate a new problem: \textbf{M}ultimodal \textbf{D}ialogue \textbf{R}esponse \textbf{G}eneration (\textbf{MDRG}), that is, given the dialogue context, the model should not only generate a pure text response but also have the capacity to generate a multimodal response (e.g., containing both image and text). We argue that there are still some hindrances to application, since (1) the sophisticated neural end-to-end architecture will overfit to very few well-annotated training data (e.g., a few existing 10k multimodal dialogues). Evidence is that when discussing the topics outside the training data domain, its performance drops dramatically; and (2) as human effort is expensive, it is not easy to collect enough training data for a new domain. Based on the above facts, we take a step further to extend the assumption of MDRG to a low-resource setting where only a few multimodal dialogues are available.

To tackle the above challenges, our key idea is to make parameters that rely on multimodal dialogues small and independent by disentangling textual response generation and image response generation, and thus we can learn the major part of the generation model from text-only dialogues and <image description, image> pairs that are much easier to be obtained. Specifically, we present  \textbf{Divter}, a novel conversational agent powered by large-scale visual world experiences. As shown in Figure \ref{fig:model}, our Divter is made up of two Transformer-based \cite{NIPS2017_3f5ee243} components: a multimodal dialogue response generator, and a text-to-image translator. Divter takes the dialogue context as input, then generates a textual sequence which may contains a text response or a textual image description or both of them. The text-to-image translator takes above image description as condition, then generates a realistic and consistent high resolution image. Both components are independent with the opposite knowledge, and thus can be pre-trained using a large number of text-only dialogues and the <image description, image> pairs respectively. The end-to-end Divter depends on the multimodal dialogues constructed as the tuple: (\begin{itshape}dialogue context, text response / <image description, image>\end{itshape}) , but the joint learning and  estimation of the two components just require a few training examples depending on specific domains.

%

Contributions of this work are three-fold:
\begin{itemize}
\setlength{\itemsep}{0pt}
\item{To the best of our knowledge, it is the first work on the multimodal dialogue response generation. We explore the task under a low-resource setting where only a few multimodal dialogues are assumed available.}
\item{We present Divter, a novel conversational agent which can effectively understand dialogue context and generate informative text and high-resolution image responses.}
\item{Extensive experiments on PhotoChat Corpus \cite{zang-etal-2021-photochat} indicate the effectiveness of Divter, it achieves a significant improvement with pure text dialogue generation model and retrieval-based image sharing method.}
\end{itemize}

\section{Related Work}

\subsection{Textual Dialogue Response Generation}
End-to-end response generation for textual open-domain dialogues is inspired by the successful application of neural sequence-to-sequence models on machine translation \cite{sutskever2014sequence}. On top of the basic architecture \cite{shang-etal-2015-neural, vinyals2015neural}, the vanilla encoder-decoder method is widely extended to address the critical challenges in open-domain dialogue systems, 
including improving the diversity of responses \cite{li-etal-2016-diversity, zhao-etal-2017-learning, ijcai2018-614}, 
modeling conversation contexts \cite{Serban_Sordoni_Bengio_Courville_Pineau_2016,xing2017hierarchical,zhang-etal-2019-recosa, zhao2020learning}, 
controlling attributes of responses \cite{see2019makes, zhou2018emotional, xu2019neural}, 
biasing responses to some specific personas \cite{li-etal-2016-persona, zhang-etal-2018-personalizing}, incorporating extra knowledge into generation \cite{dinan2018wizard, Ghazvininejad_Brockett_Chang_Dolan_Gao_Yih_Galley_2018,Kim2020Sequential, li2020zero}, and building general pre-trained agents \cite{adiwardana2020humanlike,zhang2019dialogpt, roller-etal-2021-recipes, qi2021prophetnet}. 
Different from the previous works on open-domain dialogue response generation that converse freely with plain text, our work lies in the research of multimodal response generation.


\subsection{Text-to-Image Generation}
In the research of text-to-image generation, various works have been extensively studied. \citet{mansimov16_text2image} shown the Draw generative model \cite{pmlr-v37-gregor15} could generate images from natural language descriptions. \citet{pmlr-v48-reed16} proposed a  generative adversarial network to improve the image fidelity. Then some improvement methods continue to optimize the generation architecture, such as stacked generators \cite{han2017stackgan}, attentional network \cite{Tao18attngan}, and extra knowledge \cite{objgan19}. \citet{nguyen2017plug} provided a unified probabilistic interpretation of related activation maximization methods to produce high-quality images at higher resolutions. Separately, \citet{Cho2020XLXMERT} used uniform masking with a large range of masking ratios and align the suitable pre-training datasets to the proper objectives. More recently, \citet{pmlr-v139-ramesh21a} and \citep{ding2021cogview} adopt  transformer-based methods which autoregressively model the text and image tokens as a single stream of data. For this multimodal response generation scenario, we use the textual image description to bridge above textual dialogue generation and text-to-image generation models, where the image description is the output of the former and input of the latter in a low-resource setting.



\section{Problem Formalization}\label{sec:Formalization}

Suppose that we have dataset $\mathcal{D}_S= \{(U_{i}, R_{i}) \}^{n}_{i=1}$, where  $\forall{i} \in \{1,\dots,n\}$,  $U_{i} = \{u_{i,1},\dots, u_{i,n_i}\}$ is the dialogue context  with $u_{i,j}$ the $j$-th utterance, and $R_i$ is the response regarding to $U_{i}$.  $u_{i,j}$ and $R_i$ could contain two modalities: text, and image. The goal is to learn a \textbf{generation} model $P(R|U;\theta)$ ($\theta$ denotes the parameters of the model) with $\mathcal{D}_S$.  Thus, given a new dialogue context $U$, one can generate a multimodal response $R$  following $P(R|U;\theta)$.



\section{Approach}

This section first formulates the unified tokenization method for multimodal dialogues. We then introduce the two important components in our proposed multimodal dialogue response generation model (\textbf{Divter}) under low-resource scenario, including (i) textual dialogue response generator; (ii)  text-to-image translator. Figure \ref{fig:model} shows the overall of our \textbf{Divter}.


\subsection{Multimodal Tokenization}\label{sec:Representations}

To learn a multimodal generation model, we should first model the unified  representations of both text and image. Inspired by the success of DALLE \cite{esser2020taming} and VQGAN \cite{pmlr-v139-ramesh21a}, to utilize the highly expressive transformer architecture for text-to-image generation, we need to express an image in the form of a sequence, similar to what we usually do for pure text tokenization. 

\subsubsection{Text Tokenization}

The tokenization for text is already well-studied, e.g., BPE  \cite{Gage1994ANA}. This work uses 50257 BPE-encoded tokens and distributed embedding of Transformer architecture \cite{vaswani2017attention} to model the texts in a dialogue.

\subsubsection{Image Tokenization}
The tokenizer for image  is a discrete Auto-Encoder (VQGAN\footnote{\url{https://github.com/CompVis/taming-transformers}}) $\mathcal{V}$ as shown in  Figure \ref{fig:model}. $\mathcal{V}$ uses an encoder $\mathcal{V}_{E}$ to compress each image $r^v$ of shape ${H \times W \times 3}$  into $\hat{z}$  of shape ${h \times w \times d_z}$, then each vector of dimension $d_z$ would be quantized to its closest  embedding $z_k$ in a learned, discrete codebook $\mathcal{Z} = \{ z_k\}^{K}_{k=1} \in \mathds{R}^{d_z}$ under the action of element-wise quantization  $\textbf{\textup{q}}$($\cdot$)
\begin{align}
\hspace{-0.9mm}
     z_{\textbf{\textup{q}}} =\textbf{\textup{q}} (\hat{z}) :=\left(\mathop{\arg\min}_{z_k \in \mathcal{Z}} \| \hat{z}_{ij} - z_k \|\right) \in \mathds{R}^{h \times w \times d_z} \label{eq:codebook}
\end{align}
Thus $r^v$ can be represented by a spatial collection of codebook entries $z_{\textbf{\textup{q}}} \in \mathds{R}^{h \times w \times d_z}$.  The decoder $\mathcal{V}_{D}$ maps the $z_{\textbf{\textup{q}}}$ back to a image $\hat{r^v}$ to reconstruct the input. In this work, $H = W = 256$, $h = w = 16$, $K = 16384$, $d_z = 256$. The learning details of $\mathcal{V}$ and $\mathcal{Z}$ could be found in \citet{pmlr-v139-ramesh21a}.

\subsection{Low-resource Learning Model}
\begin{figure}[!htb]
\centering
     \includegraphics[width=0.32\textwidth, scale=1, trim=260 215 420 100,clip]{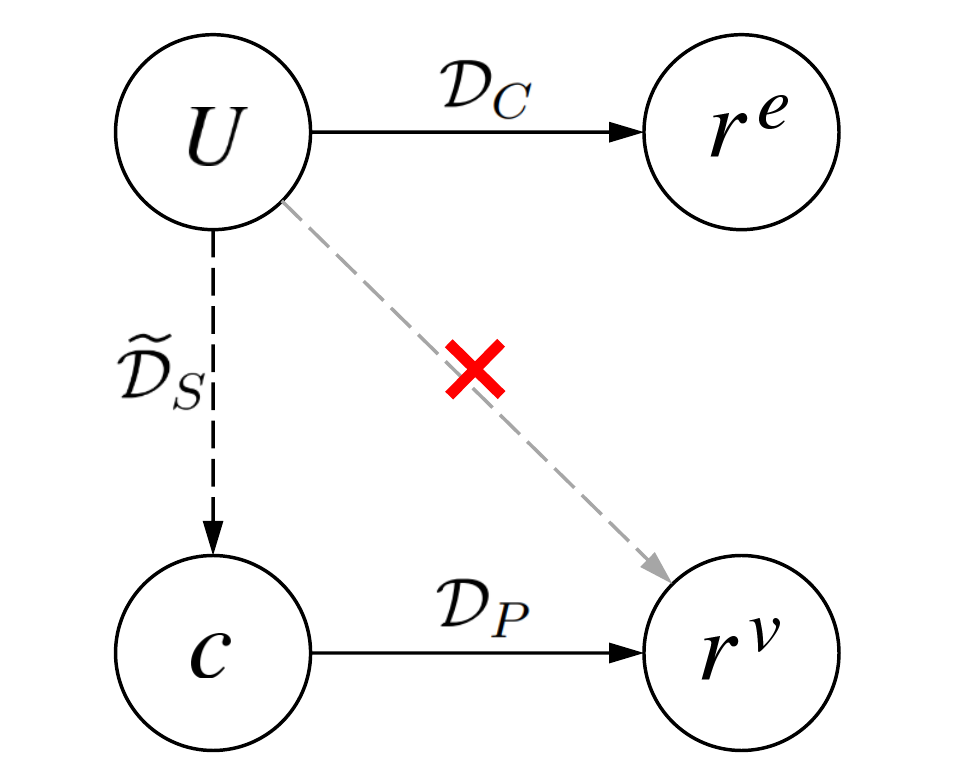}
     \caption{Abstract Logic  of the proposed approach. Solid lines mean that there exists large-scale training set to pre-train the generation model, while dotted lines mean that only very few training instances are available, ``$\times$'' means bad generation quality.}
     \label{fig:lowresource}
\end{figure}
Learning an effective multimodal generation model with a single sequence-to-sequence model often requires a large number of training instances. However, only very few multimodal dialogues are available due to the privacy restrictions on social media and the expensive human effort. On the other hand, as shown in Figure \ref{fig:lowresource}, there existed a large number of open source text-only dialogues (e.g. Reddit comments\footnote{\url{https://files.pushshift.io/reddit/}}, formulated as $\mathcal{D}_{C} = \{(U^{}_{i},  r^{e}_{i}) \}^{N}_{i=1}$ with $(U^{}_{i}, r^{e}_{i})$ a <text dialogue context, text response> pair) , and a large  number of <image description, image> pairs (e.g. YFCC100M \cite{Thomee_2016}, formulated as $\mathcal{D}_{P} = \{ (c^{}_j, r^{v}_j)\}^{M}_{j=1}$ with  $(c^{}_j, r^{v}_j)$ a <textual image-description, image> pair). Based on the above facts and the low-resource challenges on MDRG task, we adapt to incorporate generative text-to-image translation into  text-only open domain dialogue response generation. More specifically: (i) if the multimodal dialogue context contains an image, we replace the image with its description to form a text-only context, and take this context as the input of the  text-only dialogue generation model $\mathcal{G}$ (pre-trained with $\mathcal{D}_{C}$); (ii) if we need to generate an image as a part of response, we could first generation a textual description with $\mathcal{G}$, then adopt a text-to-image translator module $\mathcal{F}$ (pre-trained with $\mathcal{D}_{P}$) to translate the description to a synonymous image. To bridge $\mathcal{G}$ and $\mathcal{F}$, we further extend the formalization of $\mathcal{D}_S$ to a new $\mathcal{\widetilde{D}}_S$ in which each image $r^v$ is paired with its textual description $c$. Both the (i) and (ii) actions can be independently learned, which becomes the key to aiding the small $\mathcal{\widetilde{D}}_S$ with the large $\mathcal{D}_{C}$ and $\mathcal{D}_{P}$.


By this means, the current goal is to learn a generation model $P(R|U;\theta)$ with $\mathcal{D} = \{\mathcal{\widetilde{D}}_S, \mathcal{D}_{C}, \mathcal{D}_{P}\}$. With the pre-trained $\mathcal{G}$ and $\mathcal{F}$ available, we finally use $\mathcal{\widetilde{D}}_S$ to jointly finetune  $\mathcal{G}$ and $\mathcal{F}$ to obtain the capacity of generating multimodal responses.

Figure \ref{fig:model} illustrates the architecture of our model. The model is made up of two components: a textual dialogue response generator  $\mathcal{G}$ and a text-to-image translator $\mathcal{F}$. In the rest of this section, we will elaborate these two modules  in detail.

\subsubsection{Textual Dialogue Response Generator}

The textual dialogue response generator $\mathcal{G}$ is a sequence-to-sequence model based on the Transformer architecture \cite{vaswani2017attention}, it consists of a 24-layers Transformer with a hidden size of 1024 and 16 heads. Specifically, given a text dialogue context $U = \{u_{1},\ldots,u_{l}\}$ from $\widetilde{D}_S$ as source, and the target is a text $\widetilde{R} = \{ w_1, \cdots, [\textup{SEP}], [\textup{DST}], \cdots, [\textup{SEP}], \cdots, w_T \}$ with $w_t$ the $t$-th word, the [DST] token means the following subsequence is a textual image description $c$. The generation loss is defined by
\begin{align}
\mathcal{{L}_{G}}_{} & = \mathds{E}_{(U, \widetilde{R}) \sim \widetilde{D}_S }[-\log p(\widetilde{R})]\\
p(\widetilde{R}) & = \prod_{t} p(w_t | U, w_{1:t-1})\
\end{align}
\noindent\textbf{Inference \ }  Given a new text dialogue context $U$, when a generated image description $c$ occurs, it will be fed into the following text-to-image translator, then constructed to the codebook embeddings of its synonymous image.

\subsubsection{Text-to-Image Translator}

The text-to-image translator  $\mathcal{F}$  is also a sequence-to-sequence generation model  based on the Transformer architecture, it consists of 24-layers Transformer with a hidden size of 1024 and 16 attention heads. Given an image $r^v \in \mathds{R}^{H \times W \times 3}$ and its textual description $c = \{w_1, \cdots, w_T \}$ from $\widetilde{D}_S$, with the $\mathcal{V}_{E}$ and $\mathcal{Z}$ available, we can  represent $r^v$ in terms of the codebook indices of its encodings. More precisely, the quantized  encoding of image $r^v$ is  given by $z_{\textbf{\textup{q}}} = \textbf{\textup{q}}(\mathcal{V}_{E}(r^v))\in \mathds{R}^{h \times w \times d_z}$, and could be transferred to a sequence $s \in \{ 0,\cdots,|\mathcal{Z}|-1\}^{h \times w}$ of indices from the codebook $\mathcal{Z}$, which is obtained by replacing each code with its index in the codebook $\mathcal{Z}$
\begin{align}
  s_{i,j} = k \ \ \textup{such that} \ \ (z_{\textbf{\textup{q}}})_{i,j} = z_k
\end{align}
Then we concatenate tokenized  $c$ and $s$  to a single stream of tokens 
\begin{align}
  x = \{w_1, \cdots, w_T, [\textup{SEP}], s_{1},  \cdots, s_{h \times w} \} 
\end{align}
and train an autoregressive transformer to model the joint distribution over the text and image tokens, the generation loss is defined by 
\begin{align}
\mathcal{{L}_{F}} & = \mathds{E}_{(c,r^v) \sim \widetilde{D}_S }[-\log p(x)]\\
p(x) & = \prod_{t} p(w_t | w_{1:t-1}) \prod_{i} p(s_i | c, s_{1:i-1})\
\end{align}
\noindent\textbf{Inference \ } Given a description $c$, we leverage the text-to-image translator to generate the representations $\hat{z} = \mathcal{F}(c)\in \mathds{R}^{h \times w \times d_z}$ of its synonymous image.



\subsubsection{Learning Details}

Let us denote  $\{ {\theta}_{g}, {\theta}_{\pi}, {\theta}_{\phi} \}$ as the parameters of textual  dialogue  response  generator $\mathcal{G}$, image tokenizer $\mathcal{V}$  and  text-to-image  translator $\mathcal{F}$. In the pre-training stage, we use textual dialogues $\mathcal{D}_{C}$ to estimate ${\theta}_{g}$, use the ImageNet \cite{5206848} to estimate ${\theta}_{\pi}$, use <image description, image> pairs $\mathcal{D}_{P}$ to estimate ${\theta}_{\phi}$. Then we fix ${\theta}_{\pi}$, and jointly fine-tune ${\theta}_{g}$ and ${\theta}_{\phi}$ with $\mathcal{\widetilde{D}}_S$, thus the final objective is to minimize the integrated loss
\begin{align}
\mathcal{L} = \mathcal{{L}_{G}} + \lambda \mathcal{{L}_{F}} \label{last_eq}
\end{align}
where $\lambda$ is a hyper parameter.

\noindent\textbf{Remarks. \ } In this work, we mainly focus on integrating text and image responses generation, but our proposed approach  actually provides a recipe for a general solution to low-resource MDRG in which the target modality could be  gifs, videos, or speech sounds, etc. To do that, one only needs to modify the text-to-image translator to make it compatible with the specific modality type, then pre-train a new text-to-<target modality> translator.

\section{Experiments}

\subsection{Dataset}
To evaluate the performance of Divter, we conduct comprehensive experiments on the PhotoChat dataset released by \citet{zang-etal-2021-photochat}, which is a multimodal conversational dataset consisting of 10917 images and 12286 dialogues, each of which is paired with a user image that is shared during the conversation, and each image is paired with its text description. The dataset has been split into 10286 train, 1000 dev, and 1000 test instances. More details are described in Appendix \ref{appendix:dataset}.

\subsection{Evaluation Metrics}\label{sunsec:metircs}
We conduct evaluation with both automatic metrics and human judgements. For automatic evaluation, we focus on four aspects: (1) Image Intent Prediction, the goal of this task is to predict whether a image should be produced in the next turn for given context; (2) Text Description Generation; (3) Image Generation Quality ; (4) Text Response Generation. For (1), we follow \citet{zang-etal-2021-photochat}, which  formulates the problem as a binary classification task, and use \textbf{F1} as metric; for (2) and (4), we use \textbf{PPL}, \textbf{BLEU} \citep{papineni-etal-2002-bleu}, \textbf{Rouge} \citep{lin-2004-rouge} and \textbf{F1}; for (3) we follow \citet{pmlr-v139-ramesh21a} and use Frechet Inception Distance (\textbf{FID}) and Inception Score (\textbf{IS}).

\begin{table*}[hbt]\rmfamily\scriptsize
\renewcommand\arraystretch{1.02}
\renewcommand\tabcolsep{6.9pt}
\centering
\begin{tabular}{l | c | c c c  c | c c | c c  c    c}
\hline
\multirow{2}{*}{Models} &
\multicolumn{1}{c|}{Intent}  &
\multicolumn{4}{c|}{Image Description Generation} &
\multicolumn{2}{c|}{Image Generation} &
\multicolumn{4}{c}{Text Response Generation}  \\\cline{2-12}
& F1 &{PPL} &{B-1} & {B-2} &{Rouge}  &{FID $\downarrow$} &{IS $\uparrow$}    &{PPL} &{B-1} & {B-2}  &{Rouge}  \\
\hline
BERT-base  & 53.2$^{\ast}$ & -- & -- & -- & -- & -- &    -- & -- & --& -- & -- \\
T5-3B  & \textbf{58.9}$^{\ast}$ & -- & -- & -- & -- & -- &    -- & -- & --& -- & -- \\ \hline
S2S-TF   & 47.6 & 213.81  & 1.65 & 0.17 & 1.84 & 278.63 & 4.4  $\pm$ 0.8 &  329.43  &  3.61  & 0.40  & 3.05     \\ \hline
\textbf{Divter}   & 56.2 & \textbf{5.12} & \textbf{15.08} & \textbf{11.42} & \textbf{15.81} &  29.16 & \textbf{15.8 $\pm$ 0.6}  & 59.63 & \textbf{6.52} & \textbf{1.66} & \textbf{5.69}   \\ \hline
Divter (\begin{itshape}w/o\end{itshape} $\mathcal{G}$ pre-train)   & 47.3 & 122.56 & 1.99  & 0.23& 2.60  &29.78  & 15.5 $\pm$ 0.5    &  153.62  & 4.82& 0.53 & 3.83\\
Divter (\begin{itshape}w/o\end{itshape} $\mathcal{F}_{\phi}$ pre-train)  & 55.9 &  5.23  &15.01 & 11.20& 15.63&  262.09 & 4.9 $\pm$ 0.7 & 63.76 &  6.28  &   1.51    &   5.40 \\
Divter (\begin{itshape}w/o\end{itshape}  $\mathcal{G}$, $\mathcal{F}_{\phi}$ pre-train) & 47.1 &  128.87 & 1.75 & 0.21 & 2.38 & 254.31 & 5.2 $\pm$ 0.6 &  163.85  & 4.53   & 0.48   &  3.55  \\
Divter (\begin{itshape}w/o\end{itshape} joint learning)    & 55.6 &  5.20 &  15.00 & 11.36  & 15.73 & \textbf{29.04}  & 15.4 $\pm$ 0.6  & \textbf{59.21} &  6.47  &   1.58    &   5.63\\ 
\hline
\end{tabular}
\caption{\label{font-table}  Automatic evaluation results of Divter and baselines on the test set. (\begin{itshape}w/o\end{itshape} joint learning) means fine-tuning $\mathcal{G}$ and $\mathcal{F}_{\phi}$ respectively rather than using Eq. \ref{last_eq}. Numbers in bold mean that the improvement to the best baseline is statistically significant (t-test with $p$-value $<$ 0.01).
$^{\ast}$ reported by \citet{zang-etal-2021-photochat}.}
\label{abl:Automatic}
\end{table*}

\begin{table}[hbt]\rmfamily\scriptsize
\renewcommand\tabcolsep{4.9pt}
\renewcommand\arraystretch{1.09}
\centering
\begin{tabular}{l | c c| c c |c}
\hline
\multirow{2}{*}{Models} & Context & Text & Image  & Background   & Kappa  \\
& Coherence & Fluency & Quality   & Consistency \\
\hline
SCAN   & --  & --   & \textbf{1.95} &  \underline{0.96} &  0.65  \\
S2S-TF  & \underline{0.42}  & \underline{0.58} & 0.25& 0.20  &  0.67   \\
\textbf{Divter} & \textbf{1.59}   & \textbf{1.95} & \underline{1.83}  & \textbf{1.61}  & 0.63   \\
\hline
\end{tabular}
\caption{\label{font-table}  Human evaluation results.}
\label{abl:human}
\end{table}

\begin{table}[hbt]\rmfamily\scriptsize
\renewcommand\tabcolsep{5.1pt}
\renewcommand\arraystretch{1.03}
\centering
 \begin{tabular}{l | c c c |c}
\hline
\multirow{1}{*}{Models}&
\multicolumn{3}{c|}{Overall Improvement}   & Kappa \\ \cline{2-4}
&   W(\%) & L(\%) & T(\%) \\
\hline
Divter (pure text) vs. DialoGPT  &34.4 &35.7&  29.9 &0.64 \\
Divter vs. DialoGPT   &53.5 &27.4 & 19.1 &0.68 \\
\hline
\end{tabular}
\caption{\label{font-table}  Human evaluation results.  (W, L, T) means (Win, Lose, Tie).}
\label{abl:human_compare}
\end{table}

\begin{figure}[!htb]
\centering
     \includegraphics[width=0.3\textwidth, scale=1, trim=390 210 375 150,clip]{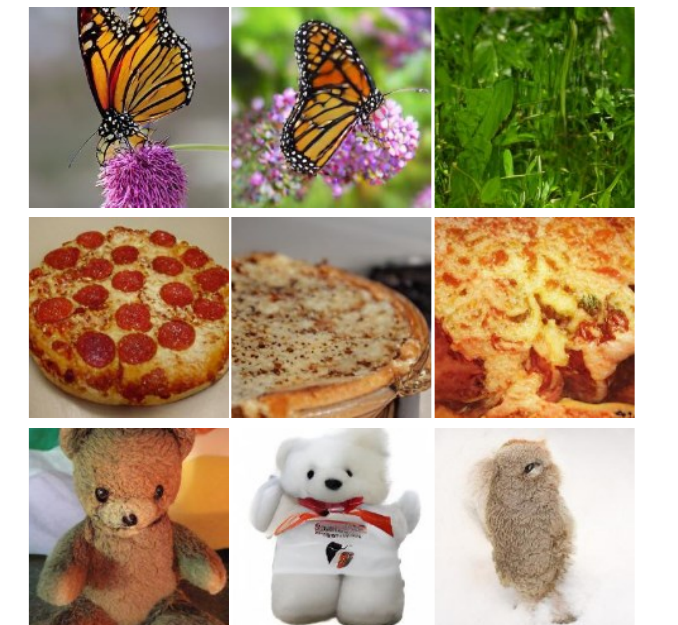}
     \caption{Qualitative assessment of various variants for image generation  with same  context as input in PhotoChat test set. 1st column: Divter. 2nd column: Divter \begin{itshape}w/o\end{itshape} $\mathcal{G}$ pre-train. 3rd column: Divter \begin{itshape}w/o\end{itshape} $\mathcal{F}$ pre-train.}
     \label{fig:Qualitative}
\end{figure}

\begin{figure*}[bht]
\centering
     \includegraphics[scale=2, width=1\textwidth,  trim=115 155 106 21,clip]{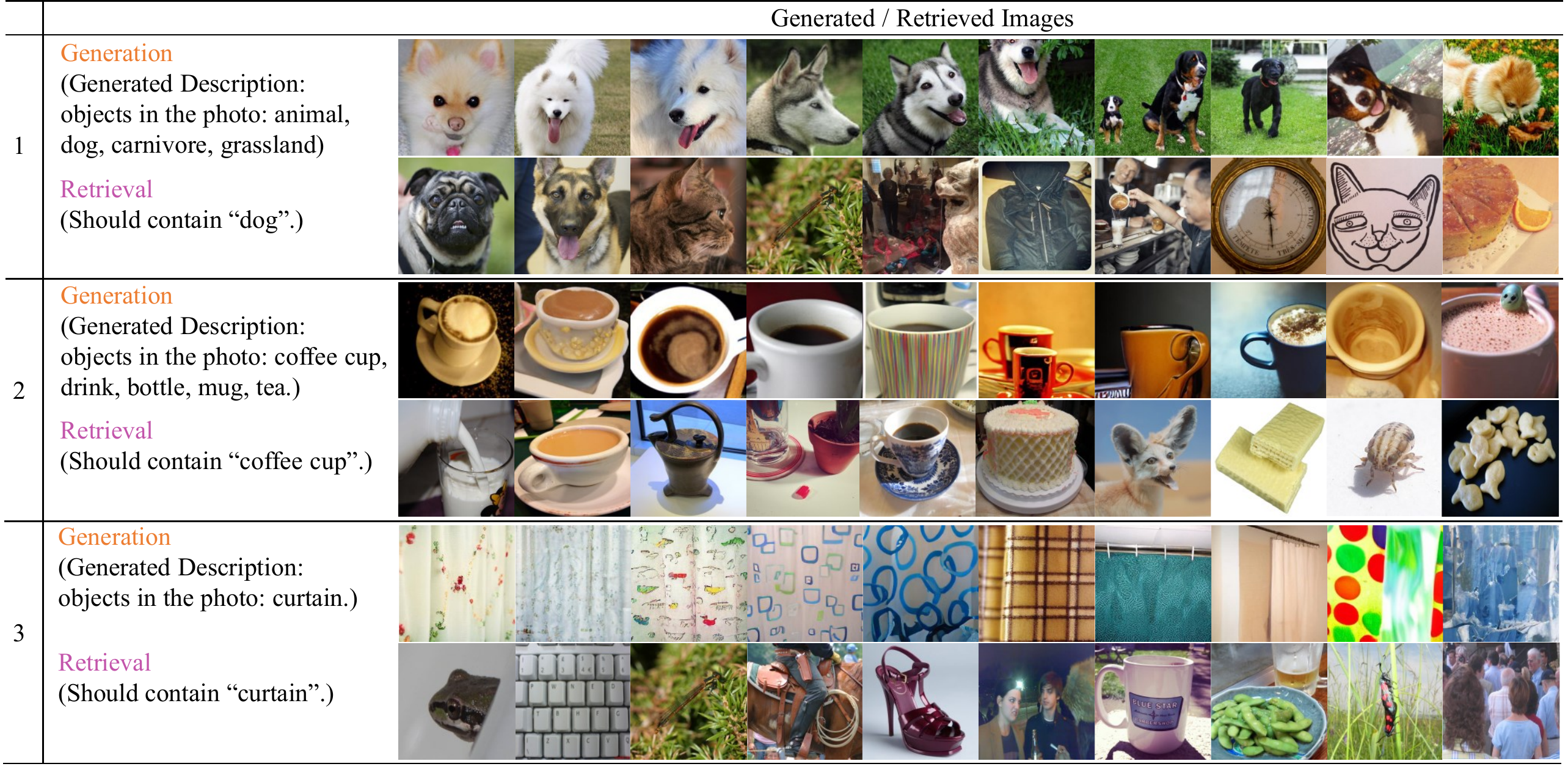}
     \caption{Examples of the images generated by Divter and the images retrieved by SCAN. The dialogue contexts are presented in Appendix \ref{appendix:contexts}.}
     \label{fig:generated_images}
\end{figure*}

For human evaluation, we randomly sample 200 dialogue contexts and generate responses from PhotoChat for Divter and baselines. Three human annotators are asked to score the response quality on a scale of \{0, 1, 2\} from four aspects: (1) \textbf{Context Coherence}: Whether the text response is coherent with the context; (2) \textbf{Text Fluency}: Whether the text response is natural and fluent;  (3) \textbf{Image Quality}: The quality (including definition and integrity) of the image response; (4) \textbf{Background Consistency of Image}: For each dialogue, We select the top-8 generated/retrieved images group and ask the annotators to decide whether the group is consistent with the dialogue background, a qualitative assessment is also shown in Figure \ref{fig:generated_images}. We report the average scores over three annotators, and the higher score means the better. 

We also compare both pure text Divter and multimodal Divter with DialoGPT, respectively. The ``pure text Divter'' means we block the [DST] token in the vocabulary in the decoding stage, so that the responses would only contain texts. We also randomly sample 200 dialogues. To each annotator, two responses from different models are presented, which are randomly shuffled to hide their sources. The annotators then judge which response is more effective in improving the dialogue experience and attractiveness. The agreement among the annotators is measured by Fleiss’ Kappa \cite{Fleiss1971MeasuringNS}.


\subsection{Implementation Details}\label{sunsec:expdetails}
For the textual dialogue response generator $\mathcal{G}$, we use DialoGPT \cite{zhang2019dialogpt} as pre-trained model initialization, trained on 147M conversation-like exchanges extracted from Reddit comment chains over a period spanning from 2005 through 2017. In the fine-tuning stage, we concatenate the context turns with the token [SEP] as a single sequence, we adopt Adam optimizer as an initial learning rate of 1e-5, and the batch size is 256, the training of PhotoChat is conducted on 16 Nvidia Tesla V100 32G GPU cards. We use beam search(size=5) to decode the text sequence. 

For the image tokenizer $\mathcal{V}$, we inherit the model released by \citet{pmlr-v139-ramesh21a}.

For the text-to-image translator $\mathcal{F}$, we randomly select 5M <categorical image description, image> pairs from ImageNet, and  <image description, image> pairs from YFCC100M \cite{Thomee_2016} as training data. We set the maximum image description length as 32, then pre-train $\mathcal{F}$ for 3.5 million steps with a batch size of 256 accumulated on 16 Nvidia Tesla V100 32G GPUs. In the fine-tuning stage, we train PhotoChat for 50000 steps. In the inference stage, we use CLIP \cite{pmlr-v139-radford21a} to rerank the generated 256 samples.


In the joint learning, we first train $\mathcal{F}$ for 48000 steps, then jointly train $\mathcal{G}$ and $\mathcal{F}$ for 2000 steps. The $\lambda$ in Eq.\ref{last_eq} is 0.2.  Early stopping on validation is adopted as a regularization strategy. All the hyper parameters are determined by grid search. More details are described in Appendix \ref{appendix:details}.

We implement the image Auto-Encoder using the code {\url{https://github.com/CompVis/taming-transformers}}, implement the Textual Dialogue Response Generator using the code {\url{https://github.com/microsoft/DialoGPT}}, and  implement the Text-to-Image Translator using the code {\url{https://github.com/lucidrains/DALLE-pytorch}}.

\begin{table*}[t]\scriptsize
\centering 
   \small
\begin{minipage}{16cm}\vspace{2mm}    \centering

   \begin{tabularx}{\linewidth}{c|c}
   \hline 
   \textbf{Example 1}  & \textbf{Example 2}\\ \hline
    \makecell[Xt]{
    \textbf{A:} OMG...the new ice cream shop is amazing. \\
    \ \ \ \ \ ......  \\
    \textbf{A:} I had the twist chocolate and vanilla but it was so fresh tasiting. like you just made it. like you just made it. \\
    \textbf{B:} I call it the malado gilato. \\
    \textbf{A:} Sam wouldn't let me have another lick bc he thought I'd eat it all. \\
   \textcolor{red}{\textbf{D:} That sounds interesting.} \\
   \textcolor{red}{\textbf{D:} Yes, could you please share it with me?} \\
    \textcolor{blue}{\textbf{D:} Objects in the photo: Chocolate Ice cream, Dairy, Drink.} \\
    \textcolor{red}{\textbf{D:}} \raisebox{-\height}[0pt][70pt]{\ \includegraphics[width=0.15\textwidth]{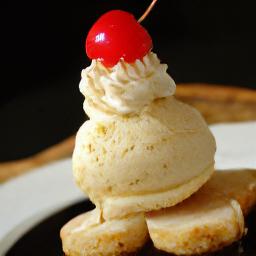}  \ \ \ \ \ \  \ \ \includegraphics[width=0.15\textwidth]{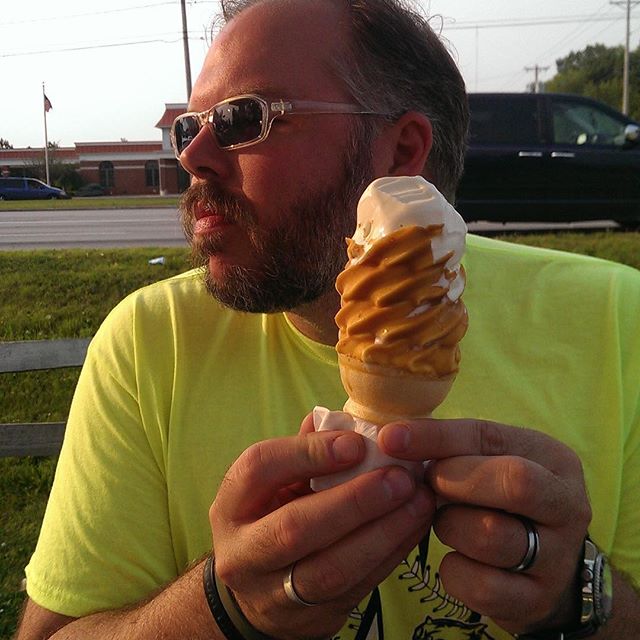} } \\
    \textcolor{red}{\textbf{D:}  Wow! The ice cream looks so delicious.} \\
    \textcolor{red}{\textbf{D:}  Sure, it tastes pretty good.}} & 
    
    \makecell[Xt]{
    \textbf{A:} Have you been out in nature lately? \\
    \textbf{B:} Yes. \\
     \ \ \ \ \ ......  \\
    \textbf{A:} I'm sitting at home now looking through some old photographs. \\
    \textbf{B:} I see. than.\\
   \textbf{A:} Would you like to see one of my favorites It's a cool shot of a honey bee near a beautiful flower. \\
    \textcolor{blue}{\textbf{D:} Objects in the photo: Honey bee, Insect, Animal, Flower.} \\
    \textcolor{red}{\textbf{D:}} \raisebox{-\height}[0pt][70pt]{\ \includegraphics[width=0.15\textwidth]{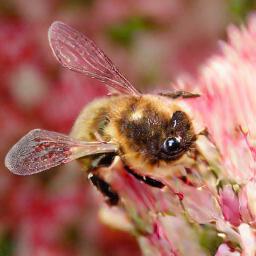} \ \ \ \  \ \  \ \ \includegraphics[width=0.226\textwidth]{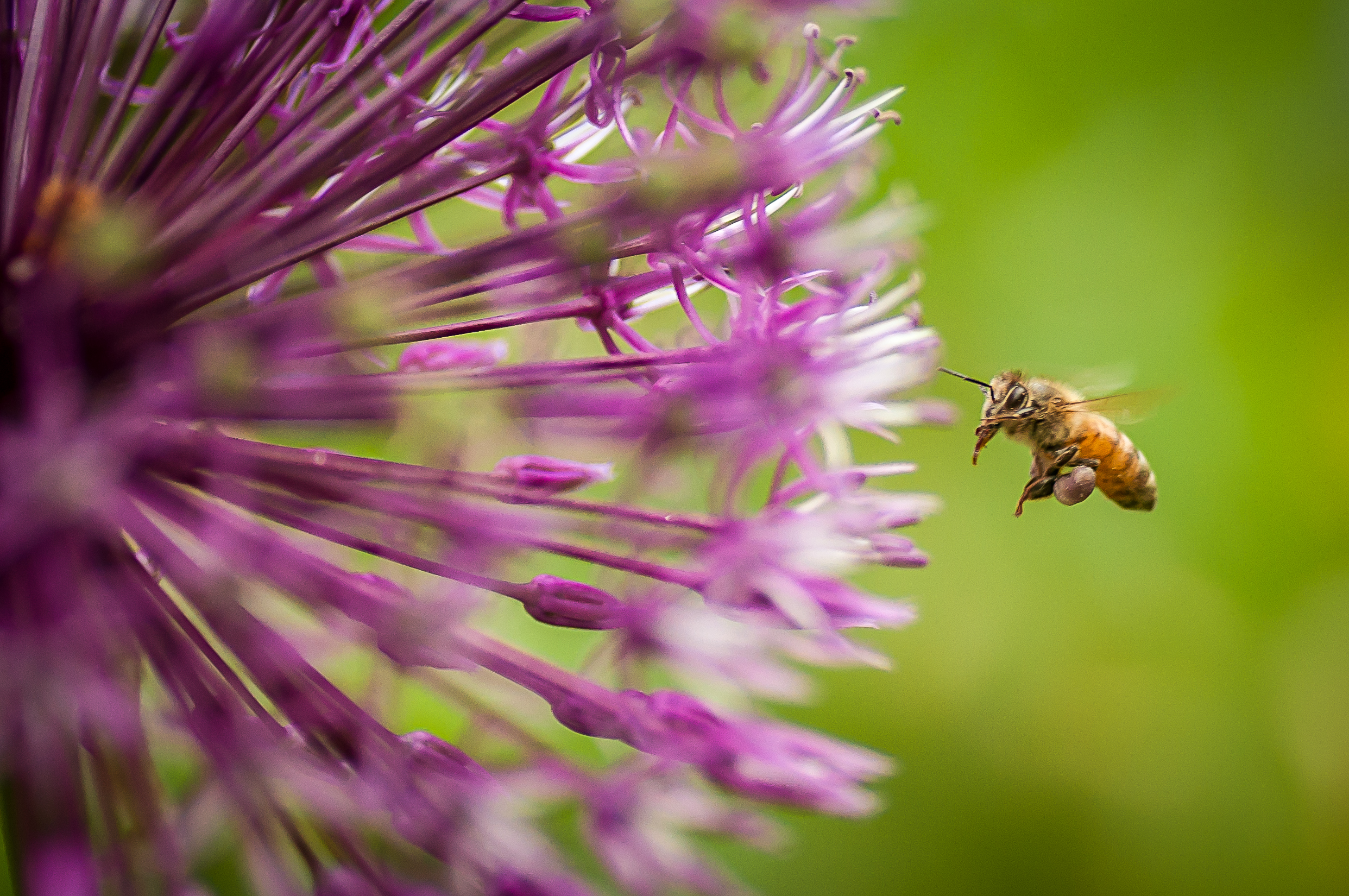} } \\
    \textcolor{red}{\textbf{D:}  It is a nice picture. Thank you for sharing.} \\
    \textcolor{red}{\textbf{D:}  Haha, just enjoy the beautiful scenery.} \\
     \textcolor{red}{\textbf{D:}  Yeah, definitely.} \\
    }\\ \hline 
    \end{tabularx}
\vspace{-1.mm}
\caption{Examples of PhotoChat test set. In each example, the turns with the prefix of ``A''/``B'' are the given context; the \textcolor{blue}{blue} text is the text description generated by Divter; the \textbf{left} image and the \textcolor{red}{red} response are generated by Divter, the \textbf{right} image is the ground-truth image.}\label{table:case_study}
\end{minipage}
\end{table*}

\subsection{Baselines}\label{sunsec:baselines}
Two pre-trained models \textbf{BERT-base} \cite{devlin-etal-2019-bert} and \textbf{T5-3B} \cite{JMLR:v21:20-074} are selected as baselines to measure the ``Image Intent Prediction'' task in Section \ref{sunsec:metircs}. They takes the text dialogue context as input, and predict ``whether a image will be shared in the next turn''.





\noindent\textbf{SCAN } is proposed by \citet{lee2018stacked}, the model captures interplay between image regions and text tokens to infer image-text similarity, SCAN achieves state-of-the-art performance of the ``Image Retrieval'' task on PhotoChat.

\noindent\textbf{S2S-TF } is a single sequence-to-sequence model with 24-layers Transformer, we only use PhotoChat to train this multimodal generation model.

\subsection{Evaluation Results}

As shown in Table \ref{abl:Automatic}, our Divter achieves not only comparable performance with the state-of-the-art retrieval-based image response intent prediction model but also achieves remarkable performance in all the generation metrics. This indicates that Divter can accurately judge the timing of generating image response with the given dialogue context, and produce text responses that are coherent to the context, and generate high-quality image responses. The significant performance gap between Divter and the baseline models (e.g. S2S-TF, Divter variants) without pre-training indicates the superiority of our proposed learning strategy. Table \ref{abl:human} reports the results of human evaluation, our Divter also significantly outperforms the baselines on most of the aspects. The comparison results shown in Table \ref{abl:human_compare} indicates (1): out Divter can achieve comparable performance on pure text response generation with DialoGPT; (2): the multimodal responses generated by Divter achieve a significant improvement on the dialogue experience and attractiveness in contrast to pure text dialogue model (DialoGPT).

\subsection{Ablation Study}

We conduct extensive ablation experiments over different variants to better understand their relative importance to the MDRG task. As shown in Table \ref{abl:Automatic}, all the variants lead to worse performance in most of the metrics. For a more intuitive comparison, the qualitative assessment results are also shown in Figure \ref{fig:Qualitative}. In particular, both quantitative and qualitative results on the ablation study validate that: (1) pre-training is crucial to low-resource multimodal dialogue response generation, since removing any component from pre-training causes performance drop when training data is small; (2) in terms of impact to performance of image generation, $\mathcal{F}$ > $\mathcal{G}$, in terms of impact to performance of text generation, $\mathcal{G}$ >  $\mathcal{F}$ ; (3) The joint learning also has contributions to Divter, indicating that leveraging the integrated learning of textual context and visual image benefits more in contrast to any single one of them.


\subsection{Case Study}

To further investigate the quality of multimodal responses generated by Divter, we show two examples on the PhotoChat test data in Table \ref{table:case_study}. The given context of the first one is about ``ice cream'', and the second one is about ``honey bee''. As we can see, Divter can not only generate a realistic high-resolution image which is coherent to the background, but also generate the informative text responses grounded on the image. Separately, The high-quality generated images are comparable to those real-world ground truths, which demonstrates the practicability of Divter.


\subsection{Discussions}


\noindent\textbf{Benefits over retrieval-based methods} To further investigate and compare the generalization capability between Divter and the retrieval-based method, we also get top-10 generated images from Divter and equivalent retrieved images from SCAN model given the same context. As shown in Figure \ref{fig:generated_images}, on the one hand, the diversity and richness of the generated images are desirable, on the other hand, those retrieved results often suffer from wrong consistency with dialogue background. For example in the second case, the dialogue is talking about ``coffee'', but the retrieved images contain some uncorrelated objects like ``milk'', ``cake'', ``dog' and ``snack''. And in the third example, all the retrieval results are mistaken since there is little ``curtain'' in the training and retrieval space. This demonstrates the fact that the performance of retrieval-based method is extremely limited in specific domains by the size of the pre-constructed conversational history repository, especially in the low-resource scenario. Furthermore, our proposed generation based method shows better generalization capability to tackle the low-resource challenge.

 
\section{Conclusion}

In this paper, we  explore multimodal dialogue response generation under a low-resource setting. To overcome the challenges from the new task and insufficient training data, we propose Divter, a neural conversational agent which  incorporates text-to-image generation into text-only  dialogue response generation, in which most parameters do not rely on the training data any more and can be estimated from large scale textual open domain dialogues and <image description, image> pairs. Extensive experiments demonstrate Divter achieves state-of-the-art results in automatic and human evaluation. In the future,  we will explore more efficient methods to inject more modalities into response generation.

\section*{Acknowledgement}
We thank anonymous reviewers for their insightful suggestions to improve this paper.



\bibliography{anthology,custom}
\bibliographystyle{acl_natbib}

\clearpage
\appendix
\section{Appendix}
\label{sec:appendix}

\subsection{Dataset}\label{appendix:dataset}
Table  \ref{table:photochat}  reports  the  statistics  of  the  PhotoChat dataset.

\begin{table}[hbt]\rmfamily
\renewcommand\arraystretch{1.02}
\renewcommand\tabcolsep{3.6pt}
\centering
\begin{tabular}{c | c | c | c | c }
\hline
Split & images   & dialogues   &  turns   &  tokens    \\
\hline
Train &  8,917 & 10,286 & 130,546 & 827,154 \\\hline
Dev &  1,000 & 1,000 & 12,701 & 80,214 \\ \hline
Test &  1,000 & 1,000 & 12,852 & 80,847 \\ \hline
Total &  10,917 & 12,286 & 156,099 & 988,215 \\
\hline
\end{tabular}
\caption{\label{font-table}  PhotoChat statistics.}
\label{table:photochat}
\end{table}

\subsection{Dialogue Contexts in Figure 5}\label{appendix:contexts}

Table \ref{table:context}  presents the textual dialogue contexts of the examples shown in the Figure \ref{fig:generated_images}.

\begin{table}[hbt]\small  
\renewcommand\arraystretch{1.12}
\centering
\begin{tabular}{p{5pt}| p{190pt} }
\hline
 & Textual Dialogue Context \\\hline
\multirow{11}{*}{1}&  A: hows your day going?  \\
& A:   beautiful sky today \\
& A: Have you been near a mountain lately     \\
& B: yes     \\
& A: beatiful right, just took a hike today with my dog.     \\
& B: my college placed in mountain area     \\
& B: super enjoy the lot     \\
& A: Oh great, do you have an aquarium at your college?     \\
& B: how is your dog     \\
& A: He is great. I'll share a pic.     \\
& B: i want to see your dog     \\
\hline
\multirow{11}{*}{2}&  B: hi  \\
& B: hello friend \\
& A: hi     \\
& A: how are you     \\
& A: i am doing well     \\
& B: how are yow     \\
& B: great     \\
& A: i am having some coffee    \\
& B: ok     \\
& A: you should come over for a cup!  \\
& A: do you like coffee?    \\
\hline

\multirow{11}{*}{3}&  A: what are you doing?  \\
& A: great moment \\
& B: Just finishing up with some work so I can start fresh tomorrow!!     \\
& B: Great moment? What's that mean? \\
& A: you chat dude    \\
& B: ???    \\
& A: Curtain you have like     \\
& B: I don't understand.    \\
& A: why dude?      \\\hline
\end{tabular}
\caption{\label{font-table} Dialogue contexts of the examples shown in the Figure \ref{fig:generated_images}.
}
\label{table:context}
\end{table}

\subsection{More Implementation Details}\label{appendix:details}
The CLIP model assigns a score based on how well the image matches the description, we use CLIP to rerank the generated 256 samples, and select the best image as the final response.  To obtain  high-quality training set, we discard the instances with the prefix of ``The photo has your * \#'' in descriptions, ``*'' includes ``mom'', ``dad'', ``daughter'', ``sister'',  ``uncle'', etc. ``\#'' is name of a person. To build the training set for text-to-image translator $\mathcal{F}$ from ImageNet, we combine the text ``Objects in the photo:'' and  textual categorical name of each image to build the <categorical image description, image> pair. To train the baseline S2S-TF model, we also use the image tokenizer  $\mathcal{V}$ to tokenize each image, and combine the image tokens with text tokens to form a single stream as the generation source/target.

\end{document}